\documentclass[conference]{IEEEtran}
\IEEEoverridecommandlockouts
\usepackage{cite}

\usepackage{amsmath,amssymb,amsfonts}
\usepackage{algorithmic}
\usepackage{graphicx}
\usepackage{textcomp}
\usepackage{xcolor}

\usepackage[ruled,vlined]{algorithm2e}
\usepackage{float}
\usepackage{array}
\usepackage{multirow}

\def\BibTeX{{\rm B\kern-.05em{\sc i\kern-.025em b}\kern-.08em
    T\kern-.1667em\lower.7ex\hbox{E}\kern-.125emX}}
\begin{document}

\title{Meta-heuristic Optimizer Inspired by the Philosophy of Yi Jing
\thanks{This work was supported in part by the Ministry of Higher Education Malaysia through the Fundamental Research Grant Scheme (FRGS/1/2023/ICT02/XMU/02/1), and Xiamen University Malaysia through Xiamen University Malaysia Research Fund (XMUMRF/2022-C10/IECE/0039 and XMUMRF/2024-C13/IECE/0049), Shenzhen College Stability Support Plan (GXWD20231128103232001), Department of Science and Technology of Guangdong (2024A1515011540), Shenzhen Start-Up Research Funds~(HA11409065), National Natural Science Foundation of China (12204130).}
}


\author{\IEEEauthorblockN{Yisheng Yang\IEEEauthorrefmark{2}, Sim Kuan Goh\IEEEauthorrefmark{2}, Qing Cai\IEEEauthorrefmark{3}, Shen Yuong Wong\IEEEauthorrefmark{2}, Ho-Kin Tang\IEEEauthorrefmark{1}}
\IEEEauthorblockA{\IEEEauthorrefmark{2}School of Electrical Engineering and Artificial Intelligence,
Xiamen University, Malaysia. \\
\IEEEauthorblockA{\IEEEauthorrefmark{3}School of Electronic Engineering, Xidian University, Xi’an, China.}
\IEEEauthorblockA{\IEEEauthorrefmark{1}School of Science, Harbin Institute of Technology~(Shenzhen).}
Corresponding: simkuan.goh@xmu.edu.my, denghaojian@hit.edu.cn}
}


\maketitle

\begin{abstract}
Drawing inspiration from the philosophy of Yi Jing, the Yin-Yang pair optimization (YYPO) algorithm has been shown to achieve competitive performance in single objective optimizations, in addition to the advantage of low time complexity when compared to other population-based meta-heuristics. Building upon a reversal concept in Yi Jing, we propose the novel Yi optimization (YI) algorithm. Specifically, we enhance the Yin-Yang pair in YYPO with a proposed Yi-point, in which we use Cauchy flight to update the solution, by implementing both the harmony and reversal concept of Yi Jing. The proposed Yi-point balances both the effort of exploration and exploitation in the optimization process. To examine YI, we use the IEEE CEC 2017 benchmarks and compare YI against the dynamical YYPO, CV1.0 optimizer, and four classical optimizers, i.e., the differential evolution, the genetic algorithm, the particle swarm optimization, and the simulated annealing. According to the experimental results, YI shows highly competitive performance while keeping the low time complexity. The results of this work have implications for enhancing a meta-heuristic optimizer using the philosophy of Yi Jing. While this work implements only certain aspects of Yi Jing, we envisage enhanced performance by incorporating other aspects. 
\end{abstract}

\begin{IEEEkeywords}
Yi Jing, Meta-heuristic Optimizer
\end{IEEEkeywords}

\section{Introduction}
In the long contest of evolution pressure in nature, living beings have developed different survival strategies. The retaining strategies being adapted are the ones that pass on generations through natural selection. They are often highly optimized as a result of million years of evolution. These amazing and brilliant strategies have been a great source of inspiration for a variety of heuristic optimization techniques~\cite{mirjalili2020genetic} and applications~\cite{du2024knowledge,jiang2023enhancing}, for examples, the cuckoo breeding behavior inspired the cuckoo search~\cite{Yang2009}, the musical performance process inspired harmony search algorithm~\cite{Al-Shaikh2021-qk}, the hunting strategy of a grey wolf pack inspired the Grey Wolf search~\cite{Ala2019}, the evolution principle of DNA inspired the Genetic Adaptive approaches~\cite{Brindle1981-gr}, machine learning~\cite{9931941,goh2021tunnel,goh2018spatio}. Among living species, human is unique, reflected by the unprecedented accumulation in cultural and philosophical contexts. These immortal gems represent the trials of humans to interpret nature. The retaining philosophies, similar to survival strategies, also passed through the contest of time. 

\begin{figure*}
\begin{centering}
    \includegraphics[width=0.85\textwidth]{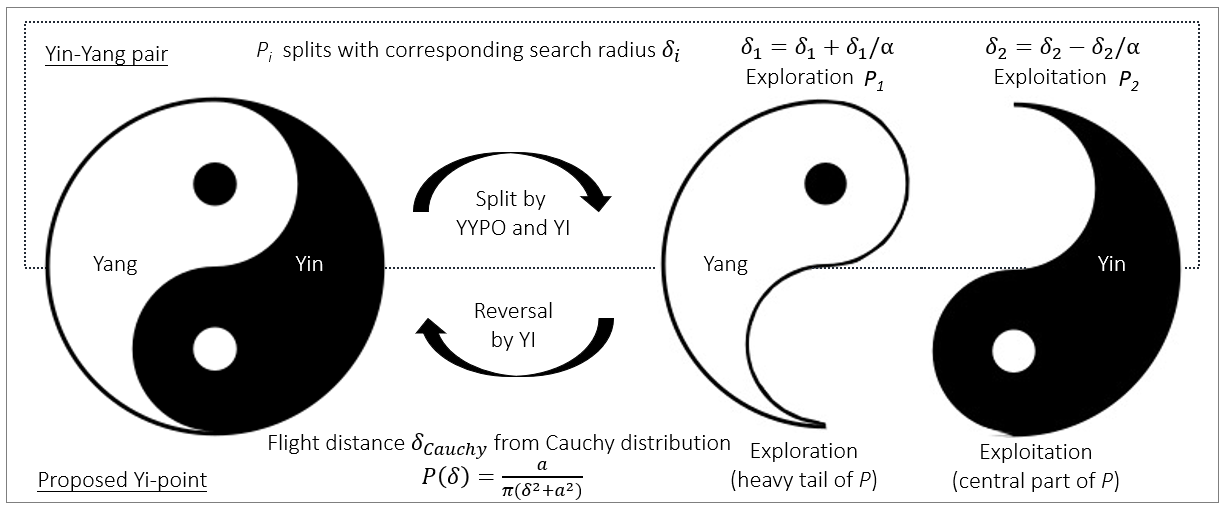}
    \par\end{centering}\vspace{-2mm}
    \caption{Illustration of the conceptual difference between the Yin-Yang pair optimization~(YYPO) and the Yi algorithm~(YI). In the Yin-Yang pair of YYPO, two points $P_1$ and $P_2$ are used to facilitate the exploration and exploitation. $P_i$ splits with the corresponding search radius $\delta_i$, where it is modified by the expansion/contraction factor $\alpha$ in the archive stage. The proposed Yi-point of YI implements the reversal and uses Cauchy flight to strike a balance between exploration and exploitation. The flight scope $\epsilon$ is divided by the decay rate $\sigma$ in the archive stage.\label{fig:fig1}}   
\end{figure*}

In ancient China, one of the famous philosophy Yi Jing~(I Ching), also known as The Book of Changes, carried the profound philosophy of harmony and unity. The Yi Jing script reads ``Taiji generates two complementary forces, two complementary forces generate four aggregates, four aggregates generate eight trigrams''. In ancient times, the ancestors tried to use the binary hierarchy in describing the phenomenon in the universe. It is interesting that the binary structure shares similarity with the Boolean algebra that modern computing relies on. The ideas in Yi Jing have good potential in inspiring new conceptual developments in algorithms.

The Yi Jing philosophy starts with a system of symbolic representation. It has been developed in many generations, resulting in rich cultural contents. The fundamental concepts include the Taiji, Yin-Yang, reversal property, rules of change in hexagrams etc. Starting with a unified object Taiji, two counterparts are generated, Yin and Yang. Usually we interpret Yin as the receptive part, and Yang as the active part. The Yin and Yang counter-balance with each other, the harmony is achieved when two forces balanced each other, which is represented in the Taiji graph. The Yin and Yang also can generate more complicated symbols including four aggregates, and eight trigrams, in which the ancient Chinese interpret them as the representations of the natural phenomenon. Different levels of interpretation are all reversal in this system~(as illustrated between Taiji and Yin-Yang in Fig.~\ref{fig:fig1}), and the dynamics of the system follows sophisticated rules of change developed.

In 2016, the Yin-Yang idea had inspired Punnathanam et al. to develop the Yin-Yang pair optimization (YYPO) \cite{Punnathanam2016a}. The algorithm uses two points to form the Yin-Yang pair to perform the optimization task. One point focuses on doing exploitation, while the other point focuses on doing exploration. YYPO is shown to possess significantly lower time complexity and better optimization performance compared to a number of algorithms. 
Later in 2017, the dynamical archive was incorporated in YYPO to formulate dynamical YYPO~(dYYPO), where its performance has a great improvement~\cite{Maharana2017} compared with YYPO. Also, Heidari et al. combined the idea of PSO and YYPO to develop PSOYPO optimizer, which gives outstanding performance in solving the uncapacitated warehouse location problem~\cite{Heidari2017}. YYPO has shown a wide application capability, e.g., to play an important role in a proposed control scheme in solar energy harvesting~\cite{Yang2009}. In 2020, 3D-YYPO is proposed to work on wind turbines design and optimization, in which a new splitting method for the three-dimensional problem is used, giving better convergence iteration number and convergence time compared to YYPO and PSO~\cite{Song2020}.


In this paper, we are inspired by another important aspect of the philosophy of Yi Jing -- the reversal property: Taiji can generate Yin and Yang, and Yin and Yang can also combine back to form Taiji. We unify the Yin-Yang pair into a single Yi-point to achieve an enhanced balance between the exploration process and the exploitation process. We proposed a highly competitive Yi algorithm~(YI), in which we replace the Yin-Yang pair with the Yi-point empowered by the heavy-tailed Cauchy flight. This paper builds upon our previous work~\cite{tang2021yi} on single objective optimization and presents an extended analysis of the proposed algorithm on higher dimensional optimization problems. The extension also includes additional comparative, complexity, and sensitivity analyses, specifically tailored to address the challenges posed by higher dimensional optimization problems.

To scrutinize the proposed YI, the benchmark from CEC 2017 is used, which contains 29 challenging functions for optimization~\cite{wu2017problem}. The proposed YI is compared against dYYPO~\cite{Maharana2017}, CV1.0~\cite{salgotra2018new}, and a few classical algorithms, i.e., DE~\cite{storn1997differential}, PSO~\cite{kennedy1995particle}, Genetic Algorithm~(GA)~\cite{Brindle1981-gr}, Simulated Annealing~(SA)~\cite{kirkpatrick1983optimization}. These classical algorithms have been the fundamental building blocks of many recent algorithms. The comparison allows us to assess YI, which builds upon the philosophy of Yi Jing, by showing Yi's relative performance to these classical algorithms that have drawn inspirations from other sources in nature originally. From the empirical results, we find that YI outperforms the classical algorithms, the dYYPO, and CV1.0 in most of the functions.




\section{Algorithm Descriptions}
In this section, we discuss the proposed YI algorithm. We first review the algorithm detail of YYPO that includes two stages, the splitting phase and the archiving phase. Then we discuss the newly proposed Yi-point, which performs the function of both exploration and exploitation, and outline the algorithm detail of YI, and give the pseudocode. 

\subsection{Review of YYPO}
The Yin-Yang pair optimization is inspired by the duality of opposite forces conflicting in nature depicted in the Yi Jing. If one could foster the balance between two forces, it results in harmony. In the field of evolutionary computing, two conflicting behaviors are exploitation and exploration. We illustrate the core idea of the algorithm in Fig.~\ref{fig:fig1}. 



YYPO algorithm starts with two randomly initialized solutions~(P1 and P2) . The fitter one of the two points is nominated as P1 and the other as P2. P1 is designed to focus on exploitation, and P2 is designed to focus on exploration. The three required parameters in terms of the minimum and maximum number of archive updates ($I_{min}$ and $I_{max}$) and the expansion/contraction factor $\alpha$ need to be specified. The number of archive updates is randomly generated between $I_{min}$ and $I_{max}$. When the iteration loop is initiated, the fitness of the two points are calculated and compared. During the updates, if P2 becomes fitter than P1, the points and their corresponding $\delta$ values are interchanged. This ensures that the loop starts with the fitter point as P1. We summarize the two main stages of YYPO as the following. The interested reader can refer to Ref.~\cite{Punnathanam2016,Punnathanam2016a} for details of YYPO.

\subsubsection{The splitting stage}
The inputs to the splitting stage are one of the points ($P_1$ or $P_2$) and its corresponding search radii ($\delta_1$ and $\delta_2$). In the splitting stage, they use two strategies that being selected randomly, namely one-way splitting and D-way splitting. Two methods are selected with equal probability in the original proposal. Later, the probabilities of choosing splitting strategy has been optimized $p_{D-way}=(\frac{D}{D+5})^2$, where the choice depends on the dimension $D$ of the problem~\cite{Punnathanam2016a}.

In the one-way splitting strategy, two solutions are generated from $P_i$  using one random number. In each new solution, only one element is changed.
\begin{eqnarray}
    P_{i,new}^j &= P_i + r\delta_i \hat{e}_j\\
    P_{i,new}^{D+j} &= P_i - r\delta_i \hat{e}_j,
\end{eqnarray}
where $r$ is a random number between 0 and 1, the superscript denotes the index of a new solution, $\hat{e}_j$ is the unit vector along $j$ direction and $j=1,2,...,D$. To generate $2D$ new solutions for both $P_1$ and $P_2$ after splitting, we need $2D$ random numbers. 

\begin{figure}[tbh]
    \centering
    \includegraphics[width=0.5\textwidth]{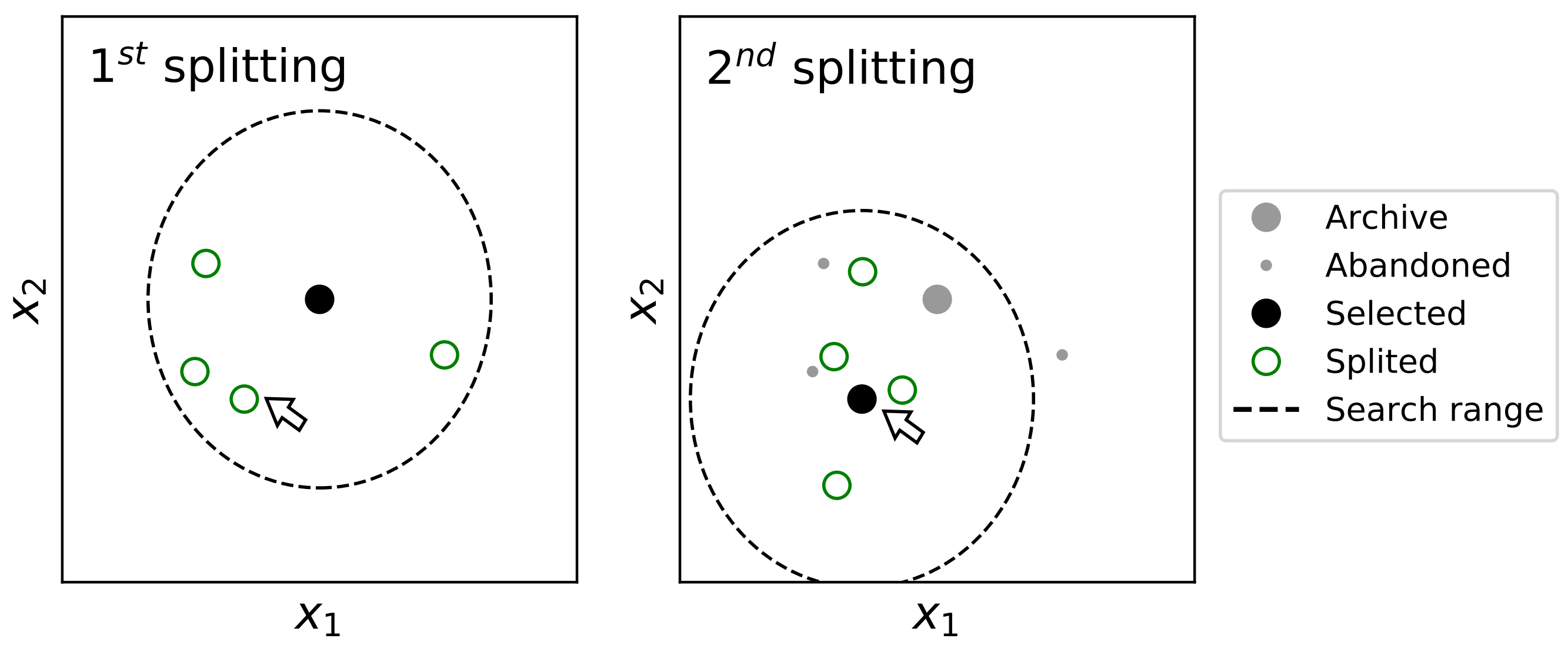}
    \vspace{-4mm}
    \caption{Schematic illustration of the splitting process. We demonstrate the D-way splitting strategy in Eq.~\ref{dway}, where the search range is $\delta_i/\sqrt{2}$ associated with $P_i$. The best among the generated solution would be picked for the seed of next splitting. The splitting process continues until reaching the archive stage.}
    \label{fig:fig2}
\end{figure}

In the D-way splitting strategy, we add a random vector $\vec{r}$ to generate new solution, in which all the elements changed. 
\begin{eqnarray}
\label{dway}
    P_{i,new} &= P_i + \vec{r} \delta_i/\sqrt{2},
\end{eqnarray}
where $\vec{r}$ is randomly distributed between -1 and 1. To generate $2D$ new solutions for both $P_1$ and $P_2$ after splitting, we need $2D\times D$ random numbers. The D-way splitting process is demonstrated in Fig.~\ref{fig:fig2}.

\subsubsection{The archive stage}
The archive stage is initiated after the required number of archive updates $I$ have been reached. The archive contains $2I$ points at this stage, corresponding to the two points $P_1$ and $P_2$ being added at each update before the splitting stage. The search radius $\delta_i$ is updated as follows,
\begin{eqnarray}
\delta_1 = \delta_1 - \delta_1/\alpha\\
\delta_2 = \delta_2 + \delta_2/\alpha
\end{eqnarray}
Initially, both $\delta_1$ and $\delta_2$ are set to be 0.5. During the updating process, the searching radii $\delta_1$ of P1 decreases to facilitate the exploitation task, while $\delta_2$ of P2 increases to enhance the exploration ability.

At the end of the archive stage, the archive matrix is set to null, and a new value for the number of archive updates $I$ is randomly generated within the given bounds, e.g., $I_{min}$ and $I_{max}$. In the latter development, dYYPO used dynamical archiving to achieve better control over the exploration and the exploitation during different periods in the search, in which $I$ increases from $I_{min}$ in the beginning phase, to $I_{max}$ at the ending phase~\cite{Maharana2017}.

\subsection{The proposed YI}
The Yi Jing promotes the concept of the infinite cycles of changes in all observing phenomena. This requires the reversal property of the Yin-Yang: Taiji can generate Yin and Yang, and Yin and Yang can also combine back to form Taiji. Inspired by the reversal property between Yin-Yang and Taiji, we combine the two points into a unified point, where we call it Yi-point. The main concept is illustrated in Fig.~\ref{fig:fig1}(b). The pseudocode of the proposed YI optimizer is given in Algorithm 1.


\begin{algorithm}
\SetAlgoLined
 Initialize the Yi-point $P_0$ randomly;\\
 Calculate the fitness of $P_0$, $f(P_0)$;\\
 Set flight scope $\epsilon=D$;\\
 Set archive duration $I = I_{max}$;\\
 Set counter $i=0$;\\
 Assign the next-interval time $T_t= T_{max}/(I_{max}-I_{min}+1)$;\\
 \While{stopping criteria}{
    \If{$i>T_t$}{
    $\epsilon= \epsilon/\sigma$;\\
    $I = I-1$;\\
    $T_t= T_{max} (I_{max}-I+1)/(I_{max}-I_{min}+1)$;\\
    }
  Cauchy splitting of $P_0$ give $P_{lbest}$;\\
  Record $P_{gbest} = P_{lbest}$ if $f(P_{lbest})<f(P_{gbest})$;\\
  Set $P_0 = P_{lbest};$\\
  $i = i+1$;\\
  \If{$i==I$}{
    Set $P_0 = P_{gbest}$;\\
    Reset counter $i=0$;\\
  }
 }
 \caption{YI algorithm}
 \label{YIALGO}
\end{algorithm}

We have three major changes in YI compared to YYPO: (1) replace the Yin-Yang pair with the Yi-point, (2) use the Cauchy flight instead of one-way splitting and D-way splitting, (3) decrease $I$ from $I_{max}$ in the beginning phase to $I_{min}$ at the ending phase. 
In what follows we provide details for the key stages of the proposed YI algorithm.

\subsubsection{The splitting stage}
We replace the Yin-Yang pair with the Yi-point, in which the update needs to strike a balance between the exploration and the exploitation. In our scheme, we use the profound Cauchy flight to perform the task. It has been found that heavy tailed functions like Levy flight or Cauchy flight successfully capture the characteristics of the flight trajectories of many animals and insects~\cite{yao1999evolutionary}, which maximize the efficiency of searching foods. The underlying reason of efficiency is the balance between exploration and exploitation. The splitting equation is given as follows,
\begin{align}
    P_{0,new} &= P_0 + \epsilon \otimes \delta_{Cauchy}(\lambda),
\label{cauchy}
\end{align}
where $\delta_{Cauchy}$ is the random number vector generated from Cauchy distribution with a stability parameter $\lambda$.


Note that $P_0$ and $P_{0,new}$ are the solutions before and after the splitting. The Cauchy distribution is a fat-tailed distribution. We choose $\lambda=1.5$ for the distribution that provides the chance of far-field exploration without losing intensified local exploitation. Also, we control the scope of the flight by $\epsilon$, initialized as the dimension $D$. It should be noted that the proposed Cauchy flight is just one of the strategies to update the Yi-point; other schemes are not yet explored.


\subsubsection{The archive stage}
YI adapts the framework of dynamical archiving in dYYPO. The archive stage is initiated after the required number of archive updates $I$ have been reached. We first set the $P_0$ as the best solution found $P_{gbest}$. Then we update the flight scope by $\epsilon = \epsilon/\sigma$.

The difference between YI and dYYPO is that in YI we decrease $I$ from $I_{max}$ in the beginning phase to $I_{min}$ at the ending phase. We divide the whole optimization process into $I_{max}-I_{min}+1$ time intervals according to the number of function evaluations done. Starting from $I_{max}$, $I$ descends by one every time when reaching the archive stage, until approaching $I_{min}$ at the end of the search. 

At the start of the search, the search process favors the strategy of exploration, so a large $I$ is preferred, allowing the Yi-point to explore the function space without the hindrance of starting from the best obtained solution. In the ending phase of searching, the exploitation becomes more important, so a smaller $I$ allows the Yi-point to focus on searching for the best solution locally.

\subsection{Summary of YI}
As can be seen from the above, YI has three user-defined parameters, $I_{min}$, $I_{max}$ and $\sigma$. $I_{min}$ and $I_{max}$ are the minimum and the maximum value of $I$ that defines the number of splitting performed before setting $P_0$ as the best recorded $P_{gbest}$, and $\sigma$ is the decay rate of the Cauchy flight scope.

We outline the pseudocode in Algorithm~\ref{YIALGO}. We first initialize the Yi-point randomly in the bounded space and calculate its fitness. We initialize the archive duration $I=I_{max}$, and the search scope $\epsilon=D$. We also define the next interval $T_t$ that gives the time for the next archive stage. 

Before reaching the archive stage, we carry out $I$ time of Cauchy splitting, given in Algorithm~\ref{cauchy} where we use Eq.~\ref{cauchy}. If the updated element of the vector is out of the boundary, we choose a random position for this element. We set $P_0$ as $P_{lbest}$, the solution with the best fitness among the splitted $P_{0, new}$. We keep using $P_{lbest}$ as $P_0$ even the fitness of $P_{lbest}$ is worse than the fitness of $P_0$. Moreover, we record $P_{lbest}$ as $P_{gbest}$ if it gives a better fitness. Here, we used the concept of gbest and lbest which first employed in PSO algorithms. gbest means the global best solution found while lbest means the local best solution found in each splitting. 

We reach the archive stage once the counter $i$ exceeds $T_t$. We rescale the scope by dividing it by $\sigma$, and decrease $I$ by one. This allows us to perform more local searches and with a shorter archive duration. To maintain the elitism of the Yi-point, we also set $P_0$ as $P_{gbest}$. The splitting stage and archive stage repeat until the stopping criteria are met. At the end of the search, we give out the best solution $P_{gbest}$ we found. 

In brief summary, the YI algorithm have a few innovation points compared to YYPO or pure Cauchy flight. First, algorithms in YYPO family employ a pair of points, here we advance to use a single point. This is advantageous when we have limitation in memory compared to other population based algorithms. Second, we use consecutive Cauchy flights in a single iteration with $I$ number of times, where $I$ deduced from $I_{max}$ to $I_{min}$ during the search. This allows more exploration in the beginning as well as more exploitation in the end of the search. Third, we adapt different search scopes in different periods of the search, for example, dividing the scope by $\sigma$ when the search proceeds to another period. This could enhance the convergence of our solution precision.


\section{Experiments on Benchmarks}
In this section, we present the experiments on the CEC 2017 benchmarks. Comparisons against several typical single objective optimization algorithms are also made.

\subsection{Benchmarks}
To validate our algorithm, we use the CEC 2017 benchmark: Single Objective Bound Constrained Real-Parameter Numerical Optimization~\cite{wu2017problem}, which comprises four main groups of functions, namely, unimodal functions~(F1, F3), simple multimodal functions~(F4-F10), hybrid functions~(F11-F20) and composition functions~(F21-F30). In total, there are 29 functions with diverse search spaces. Many of functions are a combination of representative and popular test functions. For more information of the benchmarks, please refer to refs. ~\cite{Jamil2013-xe,Wang2014-fp}.

\subsection{Optimization Algorithms}

YI belongs to a new branch of optimizer inspired by Yi Jing with the realization of reversal concept. Hence, it requires the validation of assumptions and concepts by comparing with a few classical algorithms in the purest form. To quantitatively assess our methods, we compare our method against DE~\cite{storn1997differential}, PSO~\cite{kennedy1995particle}, GA~\cite{Brindle1981-gr}, SA~\cite{kirkpatrick1983optimization}, CV1.0~\cite{salgotra2018new} and dYYPO algorithms, to examine the competitiveness of our proposed method. 

A number of studies have used $\mathrm{L\acute{e}vy}$ flight process to enhance the performance of the optimize, including the implementation of it in PSO~\cite{jensi2016enhanced}, the black hole algorithm~\cite{abdulwahab2019enhanced} and the  grey wolf algorithm~\cite{amirsadri2018levy}. CV1.0~\cite{salgotra2018new} was one of them as a variant of the cuckoo search~\cite{pavlyukevich2007levy}. The Cauchy-based global search was used in the initial phase for exploration, then the Grey wolf optimization algorithm~(GWO)~\cite{mirjalili2014grey} was utilized in the remaining phase for exploitation.

\subsection{Experimental Settings}
 
We have chosen the classical implementation of DE, PSO, GA, and SA that are provided in the python package scikit-opt. Population size of 50 is used for DE, PSO and GA. The inertia weight $w$, cognitive parameter $c1$, social parameter $c2$ are set to be 0.9, 2.0, 2.0. The $F$ and crossover rate of DE are chosen to be 0.5 and 0.001. In GA, the probability of mutation and crossover are chosen as 0.001 and 0.9. In SA, min temperature, max temperature, long of chain, cooldown time are set to be 1e-7, 100, 300, 150. These hyper-parameters are chosen to be the values recommended by scikit-opt, except for the $c1$ and $c2$ of PSO. Here, we use the values recommended by~\cite{kennedy1995particle}, which produces better optimization results for PSO. The experimental results of CV1.0 and dYYPO are extracted from their respective papers. One-tail two-sample t-test is applied to assess the relative performance of our proposed methods against CV1.0 and dYYPO.

In all the searches performed in this paper, the parameters are chosen in consideration of convergence in limited times of function evaluation. We use $I_{min}=6$, $I_{max}=15$ to divide the whole process into 11 intervals. The number of intervals cannot be too small, as this ensures the desired scope of the $\mathrm{L\acute{e}vy}$ flight at the end of the search to perform good exploitation tasks. For the initial scope and the decay rate, we choose $\epsilon_0 = D$ and $\sigma=3$. The search scope increases with the dimensions of the problem to secure the exploration coverage in the function space. 

 The source code of YI was developed in PYTHON, and it is available in GitHub repository~\cite{git}. Our test was carried out on a workstation with two Intel Xeon Platinum 8222CL 3.0GHz processors, 384 GB RAM, and Ubuntu operating system.

To assess YI, we compare it with the classic version of DE, PSO, GA and SA, as well as a $\mathrm{L\acute{e}vy}$ flight-based optimizer CV1.0, and the state-of-the-art dYYPO in YYPO family. We followed their experimental settings. Both papers used the stopping criteria of $10000\times D$ function evaluations and each benchmark problem was run 51 times. We used the same numbers for our experiments to facilitate fair comparisons.

\subsection{Results on Benchmarks of CEC 2017}
In what follows we present our experimental results on the CEC 2017 benchmarks in terms of best, worst, mean and standard deviation the error values. The error values are computed by calculating the difference expected and the desired solution. Tables 
I, II, and III from our Github repository~\cite{git} showed the statistics of our experimental results for the test problems, for dimension $D = 10,30,50$, respectively. As can be seen from $D = 10$, YI performs well on most of the benchmarks as the best errors are small, while the standard deviations are also small in most cases. With the dimension of the benchmarks increases, the difficulty for finding the global optima also increases since the searching space becomes more massive. This can be reflected by the mean values for $D = 30,50$.

We also compare YI against other algorithms on the benchmarks for 50D scenarios.
We use $+$ to signify that a baseline algorithm under
consideration is better than our proposed algorithm, $-$
to signify the opposite, and $=$ to signify either they are performing similarly without statistical significance or the comparison is irrelevant.
The comparison results are recorded in  Table~\ref{table:main}.





From the last row of Table~\ref{table:main}, the total count of the win, tie, loss (w/t/l) of CV1.0 and dYYPO against YI are shown. It is observed that the proposed algorithm is better than both CV1.0 and dYYPO the majority of the time. YI won CV1.0 for 23 out 29 functions while it won dYYPO for 21 out of 29 functions. The only irrelevant comparison was found when comparing dYYPO with YI for F3. YI was found to be much better even by comparing their mean and standard deviation; however, the standard deviation of dYYPO was too large to make statistical relevance. If dYYPO had obtained a more stable and reliable result, Yi would win dYYPO for 22 out of 29 functions.

In addition, Yi is generally found to obtain a relatively smaller standard deviation across runs (Table~\ref{table:main}), which demonstrates the stability of Yi in convergence. Despite the computational simplicity of Yi, it achieved a good balance between exploration and exploitation in the benchmark.

\begin{table*}[htb]
\vspace{-2mm}
\scriptsize
\centering
\caption{Statistical results of the proposed algorithm for 50D in comparison with
others}
\vspace{2mm}
\begin{tabular}{|l @{\hspace{1.5\tabcolsep}} l|l @{\hspace{1.5\tabcolsep}} l|l @{\hspace{1.5\tabcolsep}} l|l @{\hspace{1.5\tabcolsep}} l|l @{\hspace{1.5\tabcolsep}} l|l @{\hspace{1.5\tabcolsep}} l|l @{\hspace{1.5\tabcolsep}} l|l|}
\hline
                 &   &   & \textbf{CV1.0}    & \textbf{} & \textbf{dYYPO}  & \textbf{} & \textbf{GA}  & \textbf{} & \textbf{DE}       & \textbf{} & \textbf{PSO}      & \textbf{} & \textbf{SA}       & \textbf{YI} \\ \hline
\textbf{F1}      & mean & - & 1.00E+10          & +         & 6.60E+03      & = &  7.40E+03   & +         & \textbf{1.92E+03}          & -         & 1.21E+11          & -         & 2.79E+08          & 8.94E+03    \\
\textbf{}        & std  &   & 0.00E+00          &           & 7.10E+03      &  & 7.76E+03    &           & \textbf{1.74E+03}          &           & 3.74E+10          &           & 5.17E+07          & 7.07E+03    \\
\textbf{F3}      & mean & - & 1.95E+04          & =         & 4.70E+01      & - & 2.24E+05    & -         & 8.60E+04          & -         & 2.63E+05          & -         & 6.70E+04          & \textbf{1.23E-02}    \\
\textbf{}        & std  &   & 6.27E+03          &           & 2.30E+02      &  & 5.83E+04    &           & 9.77E+03          &           & 8.54E+04          &           & 1.60E+04          & \textbf{1.78E-03}    \\
\textbf{F4}      & mean & = & 1.16E+02          & =         & 1.40E+02      & = &  1.33E+02   & +         & \textbf{5.19E+01 }         & -         & 2.50E+04          & +         & 1.08E+02          & 1.26E+02    \\
\textbf{}        & std  &   & 6.27E+03          &           & 5.00E+01      &  &  4.16E+01   &           & \textbf{3.70E+01}          &           & 1.24E+04          &           & 5.25E+01          & 4.68E+01    \\
\textbf{F5}      & mean & - & 3.41E+02          & -         & 1.90E+02      & - &  1.94E+02   & -         & 3.16E+02          & -         & 7.20E+02          & -         & 3.91E+02          & \textbf{1.47E+02}    \\
\textbf{}        & std  &   & 8.02E+01          &           & 3.90E+01      &  &  3.50E+01   &           & 1.38E+01          &           & 1.03E+02          &           & 4.33E+01          & \textbf{3.03E+01}    \\
\textbf{F6}      & mean & - & 4.85E+01          & -         & 3.80E+00      & - &  2.43E+00   & +         & \textbf{1.14E-13}          & -         & 8.28E+01          & -         & 5.57E+01          & 2.06E-01    \\
\textbf{}        & std  &   & 4.85E+01          &           & 2.00E+00      &  &  1.03E+00   &           & \textbf{0.00E+00}          &           & 1.09E+01          &           & 1.22E+01          & 2.29E-01    \\
\textbf{F7}      & mean & - & 2.74E+02          & -         & 2.60E+02      & - &  4.31E+02   & -         & 3.70E+02          & -         & 3.70E+03          & -         & 5.40E+02          & \textbf{2.04E+02}    \\
\textbf{}        & std  &   & 7.29E+01          &           & 4.30E+01      &  & 7.50E+01    &           & 1.52E+01          &           & 5.00E+02          &           & 6.64E+01          & \textbf{3.82E+01}    \\
\textbf{F8}      & mean & - & 3.29E+02          & -         & 1.90E+02      & - &  1.86E+02   & -         & 3.16E+02          & -         & 7.42E+02          & -         & 3.95E+02          & \textbf{1.40E+02}    \\
\textbf{}        & std  &   & 7.29E+01          &           & 4.70E+01      &  &  3.76E+01   &           & 1.54E+01          &           & 1.13E+02          &           & 6.16E+01          & \textbf{2.36E+01}    \\
\textbf{F9}      & mean & - & 1.00E+04          & -         & 3.50E+03      & - &  4.59E+03   & +         & \textbf{6.02E-14}          & -         & 2.26E+04          & -         & 2.62E+04          & 8.46E+01    \\
\textbf{}        & std  &   & 2.90E+03          &           & 1.90E+03      &  &  1.77E+03   &           & \textbf{5.67E-14}          &           & 4.90E+03          &           & 5.84E+03          & 1.22E+02    \\
\textbf{F10}     & mean & - & 7.10E+03          & =         & 4.80E+03      & - &  5.36E+03   & -         & 1.16E+04          & -         & 8.86E+03          & -         & 8.21E+03          & \textbf{4.58E+03}    \\
\textbf{}        & std  &   & 5.34E+02          &           & 6.40E+02      &  & 7.65E+02      &           & 3.76E+02          &           & 1.28E+03          &           & 8.15E+02          & \textbf{7.28E+02}    \\
\textbf{F11}     & mean & = & 1.66E+02          & -         & 1.90E+02      & - &  6.87E+02   & =         & \textbf{1.57E+02}          & -         & 7.89E+03          & -         & 3.91E+02          & 1.68E+02    \\
\textbf{}        & std  &   & 3.38E+01          &           & 5.20E+01      &  & 8.50E+02    &           & \textbf{1.22E+01}          &           & 7.29E+03          &           & 8.36E+01          & 4.69E+01    \\
\textbf{F12}     & mean & - & 1.00E+10          & -         & 7.80E+06      & - &  3.91E+06   & -         & 8.50E+06          & -         & 3.30E+10          & -         & 5.72E+07          & \textbf{3.14E+06}    \\
\textbf{}        & std  &   & 0.00E+00          &           & 5.10E+06      &  &  2.03E+06   &           & 3.50E+06          &           & 1.58E+10          &           & 3.43E+07          & \textbf{1.83E+06}    \\
\textbf{F13}     & mean & - & 1.00E+10          & +         & 7.60E+03     & + & \textbf{5.01E+03}    & +         & 7.97E+03          & -         & 1.36E+10          & -         & 4.23E+05          & 6.03E+04    \\
\textbf{}        & std  &   & 0.00E+00          &           & 7.40E+03      &  & \textbf{4.79E+03}    &           & 6.22E+03          &           & 8.60E+09          &           & 2.71E+05          & 2.48E+04    \\
\textbf{F14}     & mean & + & 2.05E+02          & -         & 2.90E+04      & - & 7.25E+05    & -         & 2.37E+05          & -         & 6.01E+06          & -         & 4.89E+05          & \textbf{1.49E+04}    \\
\textbf{}        & std  &   & 2.13E+01          &           & 2.80E+04      &  &  6.75E+05   &           & 1.38E+05          &           & 6.70E+06          &           & 3.74E+05          & \textbf{1.32E+04}    \\
\textbf{F15}     & mean & - & 1.37E+09          & +         & 8.20E+03      & + & 8.50E+03    & +         & \textbf{3.08E+03}          & -         & 2.20E+09          & -         & 3.88E+04          & 2.68E+04    \\
\textbf{}        & std  &   & 3.47E+09          &           & 7.00E+03      &  &  6.09E+03   &           & \textbf{1.35E+03}          &           & 3.77E+09          &           & 2.52E+04          & 1.25E+04    \\
\textbf{F16}     & mean & - & 1.53E+03          & -         & 1.30E+03      & - &  2.03E+03   & -         & 1.80E+03          & -         & 3.66E+03          & -         & 1.67E+03          & \textbf{1.03E+03}    \\
\textbf{}        & std  &   & 2.74E+02          &           & 4.10E+02      &  & 4.16E+02    &           & 1.78E+02          &           & 8.51E+02          &           & 4.52E+02          & \textbf{2.87E+02}    \\
\textbf{F17}     & mean & - & 1.25E+03          & -         & 8.90E+02      & - &   1.51E+03            & -         & 8.47E+02          & -         & 1.11E+04          & -         & 1.43E+03          & \textbf{7.36E+02}\\
\textbf{}        & std  &   & 1.85E+02          &           & 2.70E+02      &  &   2.95E+02            &           & 1.24E+02          &           & 2.33E+04          &           & 3.56E+02          & \textbf{2.04E+02}    \\
\textbf{F18}     & mean & + & \textbf{5.21E+02}          & -         & 1.80E+05      & - & 1.88E+06    & -         & 2.00E+06          & -         & 5.15E+07          & -         & 3.11E+06          & 1.32E+05    \\
\textbf{}        & std  &   & \textbf{1.19E+02}          &           & 8.10E+04      &  & 1.45E+06    &           & 8.27E+05          &           & 8.79E+07          &           & 2.47E+06          & 7.04E+04    \\
\textbf{F19}     & mean & + & \textbf{1.73E+02}          & +         & 9.60E+03      & = &  1.60E+04    & +         & 7.55E+03          & -         & 1.14E+09          & -         & 2.08E+04          & 1.63E+04    \\
\textbf{}        & std  &   & \textbf{4.17E+02}          &           & 8.80E+03      &  &  1.08E+04   &           & 3.57E+03          &           & 1.54E+09          &           & 1.44E+04          & 1.32E+04    \\
\textbf{F20}     & mean & - & 1.05E+03          & -         & 6.50E+02      & - &  1.24E+03   & -         & 6.49E+02          & -         & 1.69E+03          & -         & 1.06E+03          & \textbf{4.82E+02}    \\
\textbf{}        & std  &   & 2.14E+02          &           & 2.80E+02      &  &  3.47E+02   &           & 1.31E+02          &           & 3.16E+02          &           & 2.72E+02          & \textbf{1.93E+02}    \\
\textbf{F21}     & mean & - & 5.41E+02          & -         & 4.00E+02      & - &  3.91E+02   & -         & 5.21E+02          & -         & 9.04E+02          & -         & 6.09E+02          & \textbf{3.42E+02}    \\
\textbf{}        & std  &   & 6.27E+01          &           & 4.00E+01      &  &  3.87E+01   &           & 1.21E+01          &           & 1.11E+02          &           & 6.97E+01          & \textbf{2.92E+01}    \\
\textbf{F22}     & mean & - & 7.33E+03          & -         & 4.80E+03      & - &  6.16E+03   & -         & 1.05E+04          & -         & 9.31E+03          & -         & 8.01E+03          & \textbf{2.41E+03}    \\
\textbf{}        & std  &   & 1.99E+03          &           & 2.00E+03      &  &  9.36E+02   &           & 2.99E+03          &           & 1.25E+03          &           & 3.22E+03          & \textbf{2.59E+03}    \\
\textbf{F23}     & mean & - & 7.74E+02          & -         & 6.50E+02      & - &  6.80E+02   & -         & 7.38E+02          & -         & 1.36E+03          & -         & 9.59E+02          & \textbf{5.62E+02}    \\
\textbf{}        & std  &   & 8.06E+01          &           & 5.70E+01      &  & 4.29E+01    &           & 1.42E+01          &           & 1.77E+02          &           & 9.08E+01          & \textbf{2.86E+01}    \\
\textbf{F24}     & mean & - & 8.32E+02          & -         & 7.30E+02      & - &  7.65E+02   & -         & 8.45E+02          & -         & 1.37E+03          & -         & 1.02E+03          & \textbf{6.50E+02}    \\
\textbf{}        & std  &   & 1.21E+01          &           & 7.40E+01      &  &  5.00E+01   &           & 1.06E+01          &           & 1.68E+02          &           & 9.47E+01          & \textbf{3.12E+01}    \\
\textbf{F25}     & mean & - & 5.43E+02          & -         & 5.20E+02      & - &  5.49E+02   & =         & 5.17E+02          & -         & 1.57E+04          & =         & \textbf{5.03E+02}          & 5.09E+02    \\
\textbf{}        & std  &   & 1.51E+01          &           & 3.00E+01      &  & 3.73E+01    &           & 3.77E+01          &           & 6.70E+03          &           & \textbf{4.88E+01}          & 3.44E+01    \\
\textbf{F26}     & mean & = & 2.48E+03          & -         & 3.40E+03      & - &  4.03E+03   & -         & 4.08E+03          & -         & 1.19E+04          & +         & \textbf{9.57E+02}          & 2.72E+03    \\
\textbf{}        & std  &   & 1.88E+03          &           & 7.80E+02      &  & 5.64E+02    &           & 1.06E+02          &           & 2.25E+03          &           & \textbf{1.24E+03}          & 3.46E+02    \\
\textbf{F27}     & mean & - & 7.38E+02          & -         & 6.70E+02      & - &  8.53E+02   & +         & 5.50E+02          & -         & 1.59E+03          & +         & \textbf{4.91E+02}          & 5.65E+02    \\
\textbf{}        & std  &   & 8.21E+01          &           & 7.30E+01      &  &  9.70E+01   &           & 1.27E+01          &           & 3.48E+02          &           & \textbf{1.35E+01}          & 4.09E+01    \\
\textbf{F28}     & mean & - & 4.94E+02          & -         & 4.80E+02      & - &  5.07E+02   & -         & 4.76E+02          & -         & 9.39E+03          & -         & 4.97E+02          & \textbf{4.67E+02}    \\
\textbf{}        & std  &   & 1.93E+01          &           & 2.50E+01      &  &  1.48E+01   &           & 2.23E+01          &           & 1.80E+03          &           & 3.03E+01          & \textbf{1.63E+01}    \\
\textbf{F29}     & mean & - & 1.69E+03          & -         & 9.80E+02      & - &  1.47E+03   & -         & 1.09E+03          & -         & 5.82E+03          & -         & 1.69E+03          & \textbf{7.40E+02}    \\
\textbf{}        & std  &   & 2.29E+02          &           & 3.10E+02      &  &  3.49E+02   &           & 1.37E+02          &           & 3.01E+03          &           & 3.98E+02          & \textbf{1.97E+02}    \\
\textbf{F30}     & mean & - & 4.64E+06          & =         & 1.50E+06      & + &  1.06E+06   & -         & 2.25E+06          & -         & 1.87E+09          & +         & \textbf{8.90E+05}          & 1.41E+06    \\
\textbf{}        & std  &   & 8.59E+06          &           & 3.20E+05      &  & 3.09E+05    &           & 3.60E+05          &           & 1.70E+09          &           & \textbf{3.98E+05}          & 2.57E+05    \\ \hline
\multicolumn{2}{|c|}{\textbf{(w,t,l)}} &  & \textbf{(3,3,23)} & \textbf{} & \textbf{(4,4,21)} & & \textbf{(3,3,23)} & \textbf{} & \textbf{(8,2,19)} & \textbf{} & \textbf{(0,0,29)} & \textbf{} & \textbf{(4,1,24)} &  \\ \hline         
\end{tabular}
\label{table:main}
\end{table*}




Supplementary Figs. 1 and 2 from our Github repository~\cite{git} demonstrate the fitness  of the three best performing algorithms from Table~\ref{table:main} with respect to the normalized number of function evaluations, when tested on the 50D benchmarks.
As can be clearly seen from Fig. 3 and Fig. 4, for the majority of the benchmarks, the red curves are above other curves at the beginning of a small number of function evaluations. However, with the increase of function evaluations, the red curves go below other curves for most of the cases. This phenomenon indicates that  YI is less greedy in achieving early convergence, but utilizes and distributes the available number of function evaluations for efficient exploration. This property allows YI to sidestep local optima, achieving superior results against the other classical algorithms studied.

\section{Algorithm Performance  Analysis}
In this section, we analyze the performance of the optimization algorithms in terms of  algorithm convergence, parameter sensitivity, algorithm similarities and differences, and algorithm complexity. 
\subsection{Performance in CEC 2017}
Generally, YI performs consistently well in different types of functions. DE performs comparably well with YI in unimodal and multi-modal functions F1-F10, while SA performs comparably well with YI in composition function F21-F30.

\begin{itemize}
    \item Convergence and handling local minima: In the unimodal functions, YI performed very well in F3 Zakharov function, converging to high precision near the global minimum in all the dimensions. Its error in means is three orders of magnitude smaller than dYYPO in 50D. The result indicates the exploitation task is done much better in YI than in dYYPO. However, YI does not outperform dYYPO in F1 Bent Cigar function, where the narrow ridge of local minimum exists; this might be due to the multi-strategy of splitting in dYYPO, enabling it performs better search along the ridge. In the multi-modal functions F4 to F10, YI outperformed both dYYPO and CV1.0 in most cases. It demonstrated the competitiveness of YI in escaping different types of local minimums. Especially in F9 $\mathrm{L\acute{e}vy}$ Function, where a huge number of local minimum presents, Yi can achieve two orders of magnitude better in error. YI is highly competitive with DE, except for the function where the ridge or the large area of plateau exist, like F1 Bent Cigar function, F4 Rosenbrock function, F6 Schaffer's Function and F9 Levy function. Also, YI outperforms both PSO, GA and SA in most of the functions.

    \item Hybrid function: The hybrid functions are complicated functions involving different basic functions for different subcomponents of the variable. The efficacy of search in these functions highly depends on the searching strategy. YI and dYYPO share many similarities in the strategy design, while CV1.0 is in a completely different realm. We expect the performance of YI and dYYPO are mostly consistent. Although YI still performs better in many cases comparing to dYYPO and CV1.0, it does not significantly outperform dYYPO in any of the hybrid functions. In F13, F15, F19, dYYPO, GA and DE is better than YI significantly. In these three hybrid functions, the F1 Bent Cigar function is involved; this could result in the weak performance of YI.
    
    \item Composition functions: YI performs consistently better in the composition function than DE, PSO, GA, dYYPO and CV1.0, in which sub-functions properties are merged. SA is competitive with YI in handling composition function, as SA optimizes to a global minimum by decreasing temperature repeatedly, which is advantageous in handling composite function in which the different functions are at different energy scales. Even without sharing algorithmic features with SA, YI performs decently in handling composition functions. This proves the YI algorithm can handle complicated function landscape very well to search the global minimum. 
    
    \item Dimensionality: When we compare the performance of YI with dYYPO~\cite{Maharana2017} in different dimensions, we find YI performs better in the higher dimensions. In 10D, YI is only slightly better than dYYPO in terms of smaller mean errors among all 29 functions. The advantage becomes more and more obvious when coming to 50D, showing the capability of Yi algorithm in complicated landscapes with high dimensions. 
    

\end{itemize}
\begin{table}[h!]
\centering
\caption{Sensitivity analysis (Compare with the case $\epsilon=3$,$I_{min}=6$ and $I_{max}=15$)}
\vspace{2mm}
\begin{tabular}{|c|c|c|c|c|c|}
\hline
\textbf{$I_{min}$} & \textbf{$I_{max}$} & \textbf{$\sigma$} & Win & Tie & Loss \\ \hline
6    & 15   & 1.5   & 0   & 1   & 28   \\
6    & 15   & 5     & 2   & 13  & 14   \\
1    & 10   & 3     & 2   & 21  & 6    \\
16   & 25   & 3     & 6   & 21  & 2    \\
6    & 25   & 3     & 0   & 21  & 8   \\ \hline
\end{tabular}
\label{parameter}
\end{table}
\subsection{Sensitivity Analysis}
We have tested different choices of hyper-parameters $\sigma$, $I_{min}$ and $I_{max}$. As listed in Table \ref{parameter}, the decay rate $\sigma$ has a great influence on the optimization result. In the searching process, either decaying too fast~($\sigma=5$) or too slow ($\sigma=1.5$) deteriorates the performance of YI. We also find that the performance of YI is not very sensitive to the change in $I_{max}$ and $I_{min}$, which control the number of intervals and the archive time of each interval in the run.
    
\subsection{Algorithm Comparison}
As YI is a newly proposed variant of YYPO, we first compare YI with other variants of YYPO. Subsequently, we compare YI with other types of algorithms. 

YI inherits YYPO's point-based feature and the dynamical archiving of dYYPO. The main difference is that we further reduce the number of points used in the searching from two to one, replacing the Yin-Yang pair with the Yi-point, this further enhances the low time complexity as well as improving the control over the exploration and exploitation. Second, instead of using both one-way splitting and D-way splitting of updating in YYPO, YI use the unified $\mathrm{L\acute{e}vy}$ flight updating. Third, contrary to dYYPO that uses ascending $I$, we use descending $I$ every time when reaching the archive stage. All these improved features contributed to the better control of the exploration and the exploitation compared to dYYPO, which is believed to performs best among the YYPO family in the general single objective optimization. We have not adapted the above changes to the population-based YYPOs, like F-YYPO and PYYPO, that are tailored to the multi-objective optimizations. This research direction is left for future works.

When compared to the PSO family, similar to PSO memory of its global best point~(gbest), YI has temporary memory of the Yi-point before reaching the archive stage. This is also common for other YYPO variants that are using the Ying-Yang pair. However, one main difference with PSO is that YI's memory does not affect the decision to search using the Cauchy flight. No velocity is calculated in YI, which is advantageous in keeping the low time complexity.

In the context of evolution strategy~(ES), YI shares some similarities the $(1,\lambda)$-ES, in which the best among $\lambda$ mutants becomes the parent of the next generation while the current parent is always disregarded. It is surprising that the reduction of the Yin-Yang pair to the Yi-point leads us to another brilliant strategy in the evolution algorithm. However, we use the heavy-tail Cauchy flight in YI to perform the exploration, instead of the crossover step in the genetic algorithms that are closely related to ES. 

YI, being a variant of YYPO, has another conceptual difference with ES, where YI emphasizes the control between exploration and exploitation. In YI, the Cauchy flight is accompanied by the dynamical archive, which allows excellent control between the exploration and the exploitation throughout different phases of the search. Given the similarity between YI and ES, it is highly motivated to adapt some ideas of ES in YI or YYPO, for instance, using the covariance matrix adaptation. 

\subsection{Complexity Analysis of the Algorithms}



Complexity of an algorithm is determined by the amount of time and space required to execute it. Analysis of the algorithm refers to the process of deriving estimates for the time and space complexity. YI inherits and further improves the low time complexity property of YYPOs. For the search with a maximum number of function evaluation $N_{eval}$, the time complexity is mainly dominated by the Cauchy flight with complexity of $O(D)$, in which the generation of a new vector requires $2D$ random number. Besides, a small amount of time complexity of $O(N_{eval})$ is spent for logical comparison of fitness. Hence, the overall time complexity of YI is $O(DN_{eval})$. 


In terms of space complexity, YI has a lower memory requirement due to the simplification of the YYPOs' archive process. The simplification removes the need to record all the lbest point before the archive stage. The required memory to run the algorithm is proportional to the dimension D, i.e. three vector of dimension D, one for gbest, one for lbest, and one for the currently splited vector. 

\section{Conclusions}

The proposed method inherits the concept of simplicity from the Yi Jing, which provides the advantage of low time complexity. This allows YI to be used in the high dimensional optimization problems. Also, we have provided a brief review of YYPOs family and analyze YI with other algorithms. The analysis allows us to place YYPOs better in the context algorithm and inspires new ideas to develop YYPOs.

According to the experimental results, our proposed methods demonstrated superior performance against CV1.0 and dYYPO on CEC 2017 benchmark that contains numerous types of challenging functions. YI outperforms CV1.0 in 23 out of 29 functions, dYYPO in 21 out of 29 functions, and DE in 19 out of 29 functions. The competitive results suggested that the proposed method achieved a good balance between exploration and exploitation. Comparing to dYYPO and CV1.0, YI is better at escaping local minima and handling complicated function landscapes in most cases. In the ridge type of function landscape like Bent Cigar function, YI performs relatively weak. This could be improved if we include other types of searching strategies, like the directional Cauchy flight as in dYYPO. The Yi-point does not limit to use Cauchy flight updating strategy. 


While this work starts a new chapter for the Yi Jing inspired optimizer, the work can be extended in numerous ways. The extension of YI from single-objective optimization to multi-objective or many objective optimizations deserves research attention. 



\bibliographystyle{IEEEtran}
\bibliography{bibliography/IEEEexample}

\end{document}